\documentclass{article}
\usepackage[colorlinks]{hyperref}
\usepackage{spconf,amsmath,graphicx,booktabs,multirow,colortbl}
\usepackage{graphicx}
\usepackage{xcolor}
\usepackage{amssymb}

\title{CMP: COMPOSABLE META PROMPT FOR SAM-BASED CROSS-DOMAIN FEW-SHOT SEGMENTATION}

\name{Shuai Chen, Fanman Meng, Chunjin Yang, Haoran Wei, Chenhao Wu, Qingbo Wu, Hongliang Li \thanks{Corresponding author: Fanman Meng.}}
\address{University of Electronic Science and Technology of China}

\begin{document}
%
\maketitle
\begin{abstract}

Cross-Domain Few-Shot Segmentation (CD-FSS) remains challenging due to limited data and domain shifts. Recent foundation models like the Segment Anything Model (SAM) have shown remarkable zero-shot generalization capability in general segmentation tasks, making it a promising solution for few-shot scenarios. However, adapting SAM to CD-FSS faces two critical challenges: reliance on manual prompt and limited cross-domain ability. Therefore, we propose the Composable Meta-Prompt (CMP) framework that introduces three key modules: (i) the Reference Complement and Transformation (RCT) module for semantic expansion, (ii) the Composable Meta-Prompt Generation (CMPG) module for automated meta-prompt synthesis, and (iii) the Frequency-Aware Interaction (FAI) module for domain discrepancy mitigation. Evaluations across four cross-domain datasets demonstrate CMP's state-of-the-art performance, achieving 71.8\% and 74.5\% mIoU in 1-shot and 5-shot scenarios respectively.

\end{abstract}
\section{Introduction}
\label{sec:intro}

The rapid development of deep learning has revolutionized semantic segmentation across numerous fields. However, the requirement for extensive labeled data remains a significant bottleneck, particularly in specialized domains~\cite{candemir2013lung,codella2019skin,demir2018deepglobe} where annotation demands expert knowledge. To this end, few-shot semantic segmentation (FSS) has emerged as a promising paradigm for segmenting novel classes with limited labeled data, demonstrating impressive capabilities in scenarios where training and testing domains are aligned (e.g., $\text{PASCAL-}5^i$~\cite{shaban2017one} and $\text{COCO-}20^i$~\cite{nguyen2019feature}.). However, when encountering significant domain shifts, such as from natural images to medical~\cite{candemir2013lung,codella2019skin} or satellite imagery~\cite{demir2018deepglobe}, existing FSS methods~\cite{tian2020prior,min2021hypercorrelation,sun2024vrp} often struggle to maintain their performance. This limitation gives rise to a more challenging task: cross-domain few-shot segmentation (CD-FSS), which must simultaneously address both limited supervision and substantial domain shifts.

The emergence of foundation models, particularly the Segment Anything Model (SAM)~\cite{kirillov2023segment}, offers new possibilities for addressing these challenges. Trained on an unprecedented scale of over 1 billion masks across diverse visual scenarios, SAM has demonstrated remarkable zero-shot generalization capabilities in general segmentation tasks. Its prompt-driven architecture and rich visual understanding make it particularly promising for CD-FSS, as it can potentially leverage its broad knowledge to bridge domain gaps. However, deploying SAM for CD-FSS faces two critical limitations. First, SAM's effectiveness heavily relies on manually crafted prompts for each test image, which introduces substantial human effort in large-scale cross-domain applications, making it operationally prohibitive for practical deployment. Second, despite its broad training distribution, SAM's performance degrades notably when encountering domains that substantially deviate from its training data. These limitations raise a fundamental question: how can we develop an automated, domain-adaptive prompting mechanism that maintains SAM's powerful segmentation capabilities while effectively bridging domain gaps?

Drawing inspiration from cognitive findings that humans achieve cross-domain generalization by integrating multiple cognitive representations~\cite{taylor2021we}, we introduce the Composable Meta-Prompt (CMP) framework, which systematically generates domain-adaptive meta-prompts through a flexible and composable mechanism that accommodates diverse segmentation references. Our framework consists of three synergistic modules designed to address the aforementioned challenges. First, the Reference Complement and Transformation (RCT) module leverages large language models to facilitate semantic expansion by identifying potential concurrent negative categories relative to the target class. Second, the Composable Meta-Prompt Generation (CMPG) module automatically synthesizes meta-prompts by integrating information from multiple sources in a composable manner, eliminating the need for manual prompt design while maintaining cross-domain adaptability. Third, the Frequency-Aware Interaction (FAI) module explores frequency-domain characteristics to address domain discrepancies, operating through two complementary mechanisms: Cross-domain Frequency Alignment (CDFA), which utilizes a memory bank of frequency statistics to align domain-specific characteristics, and Support-Query Frequency Enhancement (SQFE), which performs bidirectional amplitude spectrum interaction to reduce intra-domain variations between support and query samples. Our main contributions include:

\begin{itemize}
    \item A composable meta-prompt generation mechanism that automatically synthesizes domain-adaptive prompts by integrating multiple information sources, eliminating the need for manual prompt design while maintaining cross-domain generalization ability.
    \item A frequency-domain interaction approach that effectively bridges domain gaps through cross-domain frequency alignment and support-query in-domain frequency enhancement, providing a novel perspective on domain adaptation in CD-FSS tasks.
    \item CMP attains SOTA mIoU of 71.8\% and 74.5\% in 1-shot and 5-shot settings respectively, evaluated on four challenging cross-domain datasets (DeepGlobe, ISIC2018, Chest X-ray, and FSS-1000).

\end{itemize}

\section{Related Work}
\label{sec:related_work}

\subsection{Few-Shot Segmentation}

Few-shot semantic segmentation (FSS) tackles novel class segmentation with limited labeled examples through two main pipelines. Prototype-based methods use class-specific features from support images, evolving from simple global prototype to more sophisticated multiple prototype systems~\cite{bao2024relevant}. Matching-based methods establish pixel-level correspondences between support and query features using techniques like hypercorrelation~\cite{min2021hypercorrelation} and Correlation Distillation~\cite{peng2023hierarchical}, enabling better preservation of spatial details. However, their effectiveness in cross-domain scenarios remains challenging due to domain-specific characteristics.

\subsection{Segment Anything Model}
The Segment Anything Model (SAM)~\cite{kirillov2023segment} represents a milestone in universal image segmentation through its prompt-driven architecture. SAM has demonstrated impressive performance across various domains~\cite{10647485}. Despite its remarkable zero-shot capabilities, SAM faces two critical limitations in cross-domain applications. On one hand, the model heavily relies on manually crafted prompts for each test image, which poses significant challenges for large-scale deployments. Although recent works~\cite{sun2024vrp, zhang2024bridge} have explored automated prompting strategies through visual reference prompts, the problem remains challenging. On the other hand, SAM's performance degrades substantially when encountering domains that deviate from its training distribution. While domain-specific adaptations like ASAM~\cite{li2024asam} have been proposed to address this limitation through adversarial tuning, the adaptation remains particularly challenging in few-shot scenarios where only limited target domain data is available.

\begin{figure*}[t]
  \centering
  \includegraphics[width=0.95\linewidth]{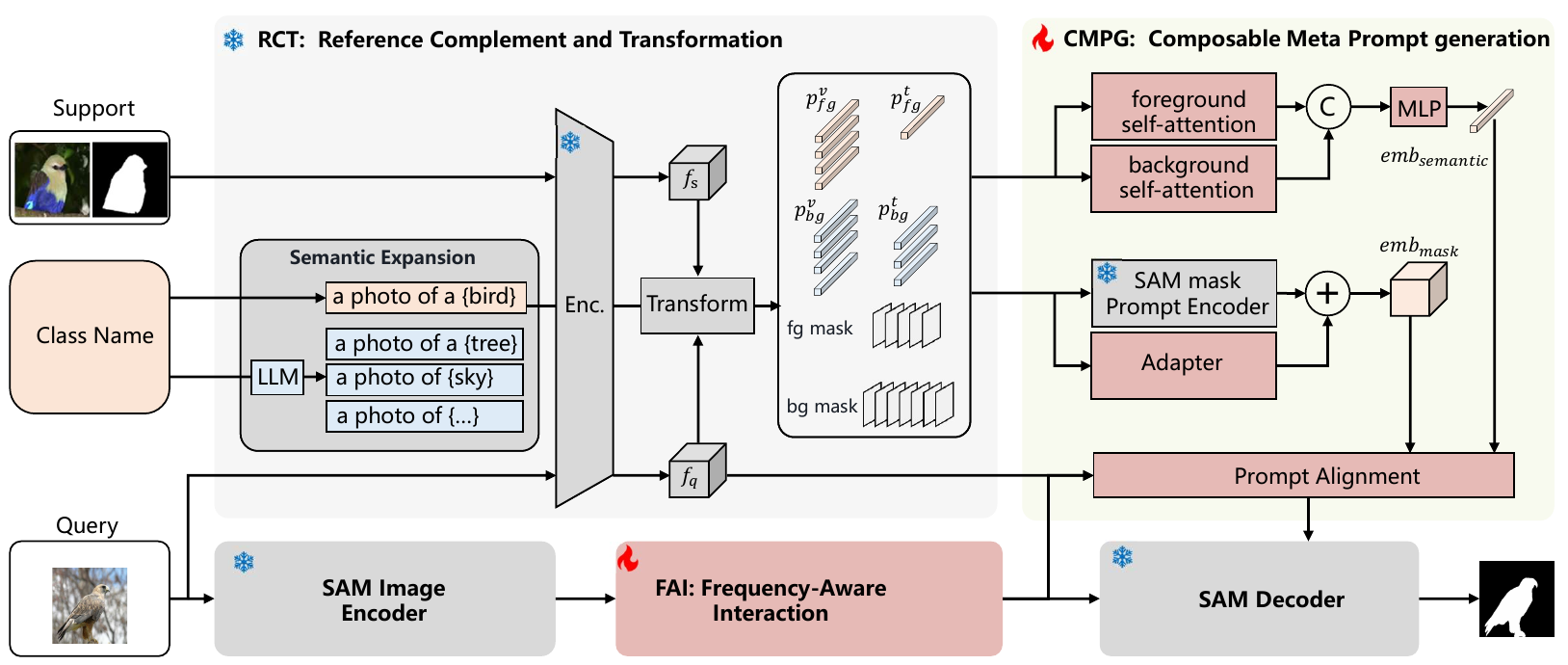}
  \caption{The Composable Meta-Prompt (CMP) framework for SAM-based Cross-Domain Few-Shot Segmentation. CMP consists of three key modules: (i) the Reference Complement and Transformation (RCT) module, (ii) the Composable Meta-Prompt Generation (CMPG) module, and (iii) the Frequency-Aware Interaction (FAI) module.}
  \label{fig:framework}
\end{figure*}

\subsection{Cross-Domain Few-Shot Segmentation}
Cross-domain few-shot segmentation (CD-FSS) tackles FSS under domain shifts. Feature transformation methods like PATNet~\cite{lei2022cross}, DMTNet~\cite{chen2024cross}, and DR-Adapter~\cite{su2024domain} focus on domain adaptation through transformation modules and style transfer. Foundation model-based approaches like APSeg~\cite{he2024apseg} leverage prompting for cross-domain generalization. However, these methods are limited by single-type prompts and spatial-only feature alignment. We address these limitations through composable meta-prompts that integrate multiple supervision signals and frequency-aware interaction.

\section{Proposed method}
\label{sec:method}

\subsection{Problem Formulation}

Cross-Domain Few-Shot Segmentation (CD-FSS) addresses the challenging scenario of transferring segmentation capabilities across distinct domains with minimal supervision. Consider a source domain $\mathcal{D}_s$ and a target domain $\mathcal{D}_t$ with different data distributions and disjoint semantic categories. Formally, let $\mathcal{D}_s = (\mathcal{X}_s, \mathcal{Y}_s)$ and $\mathcal{D}_t = (\mathcal{X}_t, \mathcal{Y}_t)$, where the domain shift is characterized by distinct input distributions ($\mathcal{X}_s \neq \mathcal{X}_t$) and non-overlapping label spaces ($\mathcal{Y}_s \cap \mathcal{Y}_t = \emptyset$). In the $K$-shot setting, the model learns from episodic samples, where each episode contains a support set $\mathcal{S} = \{(I_s^i, M_s^i)\}_{i=1}^K$ of annotated image-mask pairs and a query image $I_q$ to be segmented. The objective is to leverage meta-knowledge learned from base classes in $\mathcal{D}_s$ to effectively segment novel classes in $\mathcal{D}_t$, despite the inherent domain gap and limited target domain annotations.

\subsection{CMP Framework}

\subsubsection{Reference Complement and Transformation (RCT)}

As shown in Figure~\ref{fig:framework}, we propose the Reference Complement and Transformation (RCT) Module, which leverages semantic knowledge from large language models (LLMs) to enrich scene understanding. Specifically, we query an LLM with the prompt: \textit{``For an image containing [class name {$T_{s}$}], what other objects might co-exist?''} to generate potential co-occurring classes ${T_{\text{neg}}^j}$, which are then encoded into semantic prototypes using the text encoder $\phi_t$ of CLIP (Contrastive Language-Image Pre-training) : ${p_{fg}^t} = \phi_t(T_{s}), \quad {p_{bg}^t} = {\phi_t(T_{\text{neg}}^j)}_{j=1}^{J}$. Besides, we also extract visual prototypes from annotated regions through CLIP's visual encoder $\phi_v$:
\begin{equation}
{p_{fg}^v} = \frac{1}{|\Omega_{\text{fg}}|} \sum_{l \in \Omega_{\text{fg}}} \phi_v(I_{s}, l), \quad
{p_{bg}^v} = \frac{1}{|\Omega_{\text{bg}}|} \sum_{l \in \Omega_{\text{bg}}} \phi_v(I_{s}, l),
\end{equation}
where $\Omega_{\text{fg}}^i$ and $\Omega_{\text{bg}}^i$ denote the foreground and background regions from support mask. These prototypes are transformed into spatial priors through cosine similarity: $ \Psi_{q}^i = \text{cos}(f_{q}^i, p)$, where $p \in \{{p_{fg}^v}, {p_{bg}^v}, {p_{fg}^t}, {p_{bg}^t}\}$.

\begin{figure}[t]
  \centering
  \includegraphics[width=0.99\linewidth]{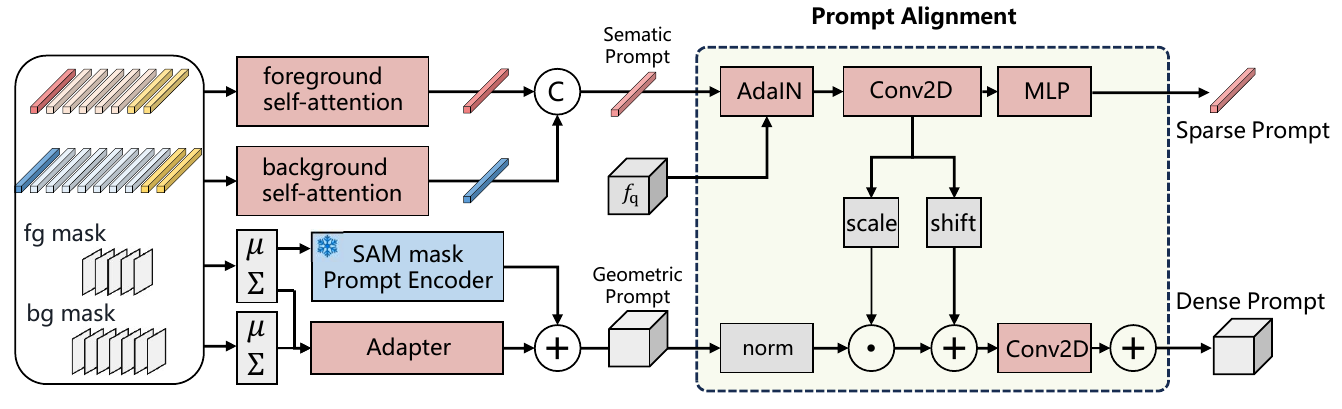}
  \caption{The Composable Meta-Prompt Generation (CMPG) module synthesizes meta-prompts by integrating information from multiple sources in a unified and composable manner, eliminating the need for manual prompt design while maintaining cross-domain adaptability.}
  \label{fig:cmpg}
\end{figure}

\subsubsection{Composable Meta-Prompt Generation (CMPG)}
The Composable Meta-Prompt Generation (CMPG) module, illustrated in Figure~\ref{fig:cmpg}, addresses SAM's reliance on manual prompts by automatically generating domain-adaptive prompts. CMPG operates at both semantic and geometric levels to generate prompts that are compatible with SAM's original prompt encoder. For semantic-level processing, we enhance foreground and background prototypes from the RCT module through self-attention. Specifically, we introduce learnable tokens ${token}_{fg}$ and ${token}_{bg}$, and compute the semantic embedding $emb_{semantic}$ as:
\begin{equation}
  \begin{aligned}
    &p_{fg}^{'} = \text{SA}({token}_{fg}; p_{fg}), \quad p_{bg}^{'} = \text{SA}({token}_{bg};p_{bg}), \\
    &{emb}_{semantic} = \mathbf{w^{T}} [{p}_{fg}^{'} ; {p}_{bg}^{'}] + \mathbf{b},
  \end{aligned}
  \end{equation}
where $\text{SA}$ denotes the self-attention mechanism, $p_{fg}^{'}$ and $p_{bg}^{'}$ are the enhanced foreground and background prototypes, and $\mathbf{w}$ and $\mathbf{b}$ are learnable parameters.

For geometric-level processing, we generate prompts that align with SAM's original prompt encoder through two complementary aspects. For mask-based prompts, considering the mismatch between SAM's binary input and our soft spatial priors $\Psi_{q}$, we introduce a learnable adapter to bridge this gap and the mask embedding $emb_{mask}$ is computed as:
\begin{equation}
  emb_{mask} = \text{MaskEncoder}({\Psi_s^i}_{i=1}^K) + \text{Adapter}({\Psi_s^i}_{i=1}^K).
  \end{equation}

Finally, we compose these prompts with query features $f_q^{sam}$ via prompt alignment block, generating the meta-prompt (dense prompt $emb_d$ and sparse prompt $emb_s$):
\begin{equation}
  emb_{d}, emb_{s} = \text{PA}(emb_{semantic}, emb_{mask}, f_q^{sam}),
\end{equation}
where $\text{PA}$ denotes the prompt alignment block that consists of convolutional blocks as shown in Figure~\ref{fig:cmpg}.

\begin{figure}[tbp]
  \centering
  \includegraphics[width=0.99\linewidth]{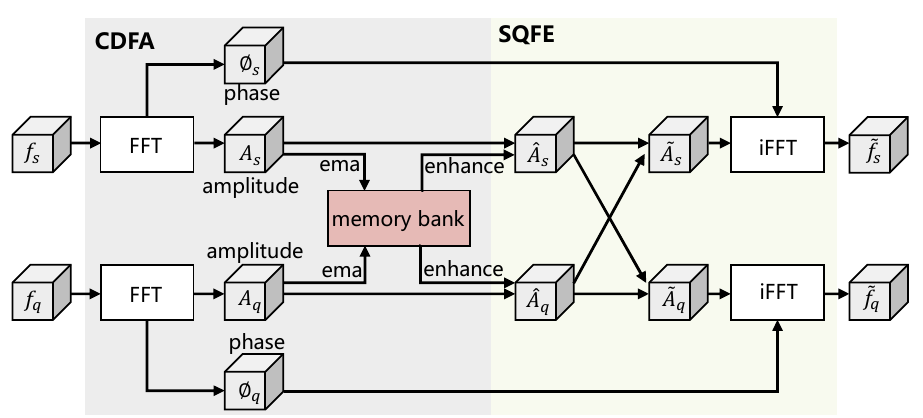}
  \caption{The Frequency-Aware Interaction (FAI) module operates through two complementary mechanisms: Cross-domain Frequency Alignment (CDFA) and Support-Query Frequency Enhancement (SQFE).}
  \label{fig:fai}
\end{figure}

\subsubsection{Frequency-Aware Interaction (FAI)}

As shown in Figure~\ref{fig:fai}, the Frequency-Aware Interaction (FAI) module consists of two complementary mechanisms: Cross-domain Frequency Alignment (CDFA) and Support-Query Frequency Enhancement (SQFE). The CDFA aligns frequency statistics between source and target domains to mitigate domain-specific characteristics, while the SQFE enhances domain-invariant features through bidirectional amplitude spectrum interaction. 

\textbf{Cross-domain Frequency Alignment (CDFA):} Given SAM's intermediate features $f = (f_s^{sam}, f_q^{sam}) \in \mathbb{R}^{C \times H \times W}$ for support and query images respectively, we first apply Fast Fourier Transform (FFT) to decompose them into amplitude and phase spectra:
\begin{equation}
\mathcal{F}(f_{s}^{sam}) = A_s e^{i\phi_s}, \quad \mathcal{F}(f_{q}^{sam}) = A_q e^{i\phi_q},
\end{equation}
where $A_s, A_q \in \mathbb{R}^{C \times H \times W}$ and $\phi_s, \phi_q \in \mathbb{R}^{C \times H \times W}$ denote the amplitude and phase spectra respectively. To facilitate cross-domain adaptation, we maintain a memory bank $\mathcal{M} \in \mathbb{R}^{T \times C}$ storing frequency statistics from source domain. For amplitude spectrum $A \in \{A_s, A_q\}$, we compute spatial-memory similarities $S \in \mathbb{R}^{H \times W \times T}$:
\begin{equation}
S_{h,w,t} = \text{cos}(A_{:,h,w}, \mathcal{M}_t) \cdot \mathbf{1}[\text{cos}(A{:,h,w}, \mathcal{M}_t) > \tau],
\end{equation}
where $\mathbf{1}[\cdot]$ is the indicator function. The memory bank is updated through weighted aggregation and exponential moving average:
\begin{equation}
\mathcal{M}_{t+1} = \alpha \mathcal{M}_t + (1-\alpha) \frac{\sum_{h,w} S_{h,w,t} A_{:,h,w}}{\sum_{h,w} S_{h,w,t} + \epsilon},
\end{equation}
where $\alpha$ is the momentum coefficient and $\epsilon$ ensures numerical stability. The enhanced amplitude spectrum is computed using the same mechanism:
\begin{equation}
\hat{A} = A + \gamma \frac{\sum_{h,w} S_{h,w,t} \mathcal{M}_{t+1}}{\sum{h,w} S_{h,w,t} + \epsilon},
\end{equation}
where $\gamma$ is a learnable scaling factor. 

\textbf{Support-Query Frequency Enhancement (SQFE):}
To strengthen domain-invariant features while maintaining semantic consistency, we perform bidirectional frequency enhancement between support and query amplitude spectra:
\begin{equation}
\begin{aligned}
\tilde{A}_q &= \text{Norm}(\hat{A}_q) \cdot (w_{s1} \hat{A}_s + b_{s1}) + (w_{s2} \hat{A}_s + b_{s2}), \\
\tilde{A}_s &= \text{Norm}(\hat{A}_s) \cdot (w_{q1} \hat{A}_q + b_{q1}) + (w_{q2} \hat{A}_q + b_{q2}),
\end{aligned}
\end{equation}
where $w_{s1}, w_{s2}, b_{s1}, b_{s2}, w_{q1}, w_{q2}, b_{q1}, b_{q2}$ are learnable parameters. The interacted features are reconstructed through inverse FFT:
\begin{equation}
  \tilde{f}_q = \mathcal{F}^{-1}(\tilde{A}_q e^{i\phi_q}), \quad \tilde{f}_s = \mathcal{F}^{-1}(\tilde{A}_s e^{i\phi_s}).
\end{equation}

\begin{table*}[htbp]
  \centering
  \caption{Performance comparison of CD-FSS Methods Using mIoU (\%) Metric. Results marked in \textbf{bold} and \underline{underlined} indicate first and second-best performance respectively. $*$ denotes results from ~\cite{chen2024cross}.}
  \label{tab:cd-fss}
  \scalebox{0.7}{
  \begin{tabular}{l|c|c|cc|cc|cc|cc|cc}
      \toprule[1pt]
      \multicolumn{13}{c}{Source Domain: Pascal VOC 2012 $\to$ Target Domain: Below}\\\hline
      \multirow{2}*{Methods}& \multirow{2}*{Publication}& \multirow{2}*{Backbone}& \multicolumn{2}{c|}{Deepglobe}& \multicolumn{2}{c|}{ISIC}& \multicolumn{2}{c|}{Chest X-Ray}& \multicolumn{2}{c|}{FSS-1000}& \multicolumn{2}{c}{\textbf{Average}}\\\cline{4-13}
      ~& ~& ~& 1-shot& 5-shot& 1-shot& 5-shot& 1-shot& 5-shot& 1-shot& 5-shot& 1-shot& 5-shot\\
      \midrule[1pt]
      PFENet*~\cite{tian2020prior}&TPAMI'20& \multirow{7}*{Res-50}& 16.9& 18.0& 23.5& 23.8& 27.2& 27.6& 70.9& 70.5& 34.6& 35.0\\
      HSNet*~\cite{min2021hypercorrelation} &ICCV'21&~ & 29.7& 35.1& 31.2& 35.1& 51.9& 54.4& 77.5& 81.0& 47.6& 51.4\\
      PATNet~\cite{lei2022cross}&ECCV'22& ~& 37.9& 43.0& 41.2& 53.6& 66.6& 70.2& 78.6& 81.2& 56.1& 62.0\\  
      HDMNet~\cite{peng2023hierarchical} &CVPR'23& ~& 25.4& 39.1& 33.0& 35.0& 30.6& 31.3& 75.1& 78.6& 41.0& 46.0\\
      DMTNet~\cite{chen2024cross}&IJCAI'24 &~& 40.1& \underline{51.2}& 43.6& 52.3& 73.7& 77.3& \underline{81.5}& \underline{83.3}& 59.7& \underline{66.0}\\
      DR-Adapter~\cite{su2024domain}&CVPR'24& ~& 41.3& 50.1& 40.8& 48.9& 82.4& 82.3& 79.1& 80.4& 60.9& 65.4\\
      ABCDFSS~\cite{herzog2024adapt}&CVPR'24& ~& 42.6& 49.0& \underline{45.7}& 53.3& 79.8& 81.4& 74.6& 76.2& 60.7& 65.0\\

      \midrule[1pt]
      HQ-SAM~\cite{ke2024segment}&NeurIPS'23 & \multirow{5}*{SAM}& 24.7& 26.8& 40.4& 47.6& 28.8& 30.1& 79.0& 81.0& 43.2& 46.2\\
      PerSAM~\cite{zhang2023personalize} &ICLR'24& ~& 36.1& 40.7& 23.3& 25.3& 30.0& 30.1& 60.9& 66.5& 37.6& 40.6\\
      Matcher~\cite{liu2023matcher} &ICLR'24& ~& \underline{48.1}& 50.9& 38.6& 35.0&- &- &- &- &- &-\\
      APSeg~\cite{he2024apseg}&CVPR'24& ~ & 35.9& 40.0&45.4&\underline{54.0}&\underline{84.1}& \underline{84.5}&79.7&81.9&\underline{61.3}&65.1\\
      \cellcolor [HTML]{EFEFEF} CMP (Ours) &\cellcolor [HTML]{EFEFEF} - & \cellcolor [HTML]{EFEFEF}~ & \cellcolor [HTML]{EFEFEF}\textbf{49.2}& \cellcolor [HTML]{EFEFEF}\textbf{52.7}& \cellcolor [HTML]{EFEFEF} \textbf{60.5}& \cellcolor [HTML]{EFEFEF} \textbf{62.6}& \cellcolor [HTML]{EFEFEF} \textbf{86.7}& \cellcolor [HTML]{EFEFEF} \textbf{88.1}& \cellcolor [HTML]{EFEFEF} \textbf{90.9}& \cellcolor [HTML]{EFEFEF} \textbf{94.6}& \cellcolor [HTML]{EFEFEF} \textbf{71.8}& \cellcolor [HTML]{EFEFEF} \textbf{74.5}\\
      \bottomrule[1pt]
  \end{tabular}}
\end{table*}

\section{Experiments}
\label{sec:experiments}

\subsection{Datasets and Evaluation Metrics}
\label{ssec:datasets}

We train our model on PASCAL VOC 2012 with SBD augmentation and evaluate across diverse target domains. The target domains span natural objects (FSS-1000~\cite{li2020fss}), remote sensing imagery (DeepGlobe~\cite{demir2018deepglobe}), and medical imaging (ISIC2018~\cite{codella2019skin} for skin lesions and Chest X-ray~\cite{candemir2013lung} for pulmonary screening). Performance is measured using mean Intersection over Union (mIoU).

\begin{table}[htbp]
  \centering
  \caption{Ablation study on DeepGlobe dataset. $\triangledown$ indicates performance drop compared to the full model.}
  \label{tab:ablation}
  \begin{tabular}{l|c|c}
  \hline
  Method & mIoU  & $\triangledown$ \\
  \hline
  CMP (Full) & 49.2 & - \\
  w/o CMPG & 42.5 & -6.7 \\
  w/o  SE in RCT & 45.2 & -4.0 \\
  w/o whole FAI & 46.0 & -3.2 \\
  w/o CDFA in FAI & 48.2 & -1.0 \\
  w/o SQFE in FAI & 46.8 & -2.4 \\

  \hline
  \end{tabular}
  \end{table}

\subsection{Implementation Details}
\label{ssec:implementation}

Our framework retains SAM's native 1024×1024 resolution with frozen parameters during training. It employs a two-stage process: meta-training on the source domain and fine-tuning on the target domain. Both stages utilize the Adam optimizer with a learning rate of $1 \times 10^{-4}$ and a batch size of 2 per GPU across four NVIDIA RTX 3090 GPUs. Source training on PASCAL VOC is conducted for 10 epochs with data augmentations, including random flipping, color jittering, and affine transformations. For target fine-tuning, we augment the limited support samples to generate pseudo queries for each episode. The entire framework is implemented in PyTorch. We adopt a combined loss function that balances between binary cross-entropy $\mathcal{L}_{bce}$ and Dice loss $\mathcal{L}_{dice}$ to optimize our model:
\begin{equation}
\mathcal{L} = \lambda \mathcal{L}_{bce} + (1-\lambda) \mathcal{L}_{dice},
\end{equation}
where $\lambda$ is the weighting coefficient, set to 0.5 for default.

\subsection{Comparison with State-of-the-Art Methods on CD-FSS Tasks}
\label{ssec:results_cd-fss}

We compare CMP with two categories of methods: ResNet-based approaches and SAM-based methods. The ResNet-based methods include recent SOTA frameworks such as DMTNet~\cite{chen2024cross}, DR-Adapter~\cite{su2024domain}, and ABCDFSS~\cite{herzog2024adapt}. For SAM-based methods, we compare with HQ-SAM~\cite{ke2024segment}, PerSAM~\cite{zhang2023personalize}, Matcher~\cite{liu2023matcher}, and APSeg~\cite{he2024apseg}. As shown in Table~\ref{tab:cd-fss}, CMP consistently outperforms all existing methods across different domains and shot settings. Specifically, CMP achieves significant improvements in challenging scenarios, with an average mIoU of 71.8\% and 74.5\% in 1-shot and 5-shot settings respectively, surpassing the previous best results by 10.5\% and 8.5\%.  On the ISIC dataset, CMP achieves 60.5\% mIoU in the 1-shot setting.

\subsection{Ablation Studies}
\label{ssec:ablation}

Table~\ref{tab:ablation} present the ablations of CMP on DeepGlobe dataset:

\textbf{CMPG and RCT module:} Removal of the Composable Meta-Prompt Generation (CMPG) module yielded the most significant performance decline (-6.7\%). Subsequently, eliminating the semantic expansion component from the RCT produced the second largest performance reduction (-4.0\%).

\textbf{Frequency-Aware Interaction:} The FAI module consists of two key components: Cross-Domain Frequency Alignment (CDFA) and Support-Query Feature Enhancement (SQFE). Removing the entire FAI module (both CDFA and SQFE) results in a significant performance drop (-3.2\%), demonstrating its crucial role in cross-domain adaptation. Removing SQFE alone leads to a -2.4\% decrease in performance, while removing CDFA causes a -1.0\% drop.

\subsection{Limitations}

The performance degrades under extreme domain shifts and when coarse foreground mask is extremely inaccurate (e.g., due to support-query mismatches).

\section{Conclusion}
\label{sec:conclusion}

We propose the Composable Meta-Prompt (CMP) framework for Cross-Domain Few-Shot Segmentation, which utilizes composable meta prompts with cross-domain frequency strategies to address prompt design and domain adaptation challenges. Built upon SAM architecture, our framework achieves SOTA performance on CD-FSS benchmarks.

\bibliographystyle{IEEEbib}
\bibliography{refs_simplified}

\end{document}